\documentclass[11pt,a4paper]{article}
\usepackage[english]{babel}
\usepackage{url}
\usepackage[utf8]{inputenc}
\usepackage{xcolor}
\usepackage{hyperref}
\usepackage{booktabs}
\usepackage{booktabs}
\usepackage{authblk}
\usepackage{xspace}
\usepackage{longtable}
\usepackage{amssymb,amsmath}
\usepackage{bm}
\usepackage{amsfonts}
\usepackage{multirow}
\usepackage[normalem]{ulem}
\usepackage{wrapfig}
\usepackage{paralist}
\usepackage{times}
\usepackage{latexsym}
\usepackage{graphicx}

\title{ConvAI3: Generating Clarifying Questions for Open-Domain Dialogue Systems (ClariQ)}

\begin{document}

\author[1]{Mohammad Aliannejadi}
\author[2]{Julia Kiseleva}
\author[3]{Aleksandr Chuklin}
\author[4]{Jeff Dalton}
\author[5]{Mikhail Burtsev}

\affil[1]{University of Amsterdam, Amsterdam, The Netherlands,
    {\tt m.aliannejadi@uva.nl}}
\affil[2]{Microsoft Research AI, Seattle, USA,
    {\tt julia.kiseleva@microsoft.com}}
\affil[3]{Google Research, Z\"{u}rich, Switzerland,
    {\tt chuklin@google.com}}
\affil[4]{University of Glasgow, Glasgow, UK,
    {\tt jeff.dalton@glasgow.ac.uk}}
\affil[5]{MIPT, Moscow, Russia,
  {\tt burtcev.ms@mipt.ru}}

\maketitle

\section{Introduction}\label{sec:intro}
This document presents a detailed description of the challenge on clarifying questions for dialogue systems (ClariQ) \emph{[pronounce as Claire-ee-que]}. The challenge is organized as part of the Conversational AI challenge series (ConvAI3)\footnote{\url{http://convai.io/}} at Search-oriented Conversational AI (SCAI) EMNLP workshop in 2020.\footnote{\url{https://scai.info}}.
The main aim of the conversational systems is to return an appropriate answer in response to the user requests. However, some user requests might be ambiguous. In IR settings such a situation is handled mainly thought the diversification of search result page~\cite{radlinski2006improving}.
It is however much more challenging in dialogue settings. Hence, we aim to study the following situation for dialogue settings: 

\begin{itemize}
    \item a user is asking an ambiguous question (where ambiguous question is a question to which one can return $>1$ possible answers);
    \item the system must identify that the question is ambiguous, and, instead of trying to answer it directly, ask a good clarifying question.
\end{itemize}

The main research questions we aim to answer as part of the challenge are the following:
\begin{itemize}
    \item RQ1: When to ask clarifying questions during dialogues?
    \item RQ2: How to generate the clarifying questions?
\end{itemize}

The rest of the document is organized as follows: Section~\ref{sec:rel_work} describes previous efforts and datasets related to the clarifying questions.
We outline the design of our ClariQ challenge in Section~\ref{sec:design}.

\section{Related work}
\label{sec:rel_work}
There were a number of attempts to study clarifying questions recently. \\ \cite{braslavski2017you} and \cite{rao2018learning} studied how users of Stack Exchange. \cite{braslavski2017you} focused on characteristics, forms, and general patterns of clarifying questions. \cite{rao2018learning} designed a model to rank a candidate set of clarification questions by their usefulness to the given post at Stack Exchange. As result, the dataset extracted from Stack Exchange was released, but it covers specific narrow topics.  Therefore, we are limited to understand clarifying questions for open-domain conversations.

Information retrieval (IR) community recently has payed close attention to the problem of generating clarifying questions in open-domain settings \\ \cite{rosset2020leading}, \cite{AliannejadiSigir19}, and \cite{zamani2020generating}. \\ Where the general settings are the following: (1) a user is issuing a keyword query, which is ambiguous, and (2) a search engine's goal is to suggest a conversational clarifying question to help to find the required information. \cite{AliannejadiSigir19} shared a \emph{Qulac} dataset with the community which allows follow-up studies. As a follow-up \cite{zamani2020generating} released a \emph{MIMICS} dataset \cite{zamani2020mimics}, which consists of queries, issued by real users, and behavioral signals such as clicks.  

ClariQ challenges is closely related to Q{\&}A domain~\cite{kwiatkowski2019natural}. For example, \cite{trienes2019identifying} made an attempt to understand an unclear question. Hence, we can assume that some solution for ClariQ might benefit from utilizing some techniques from Q{\&}A and adapted to conversational settings. 

Therefore, we can conclude that understanding and generating clarification questions has also been recognized as a major component in conversational information-seeking systems. There were a number of efforts of sharing datasets with the community to facilitate further research in this direction. Hence, our efforts to set-up challenge ClariQ on generating clarifying questions in dialogue settings are timely.

\section{Challenge Design}
\label{sec:design}
The ClariQ challenge is run in two stages. At Stage~1 (described in Section~\ref{sec:stage_1}) participants are provided static datasets consisting mainly of an initial user request, clarifying question and user answer, which is suitable for initial training, validating and testing.  At Stage~2 (described in Section~\ref{sec:stage_2}), we bring a human in the loop. Namely, the TOP-N systems, resulted from Stage~1, are exposed to the real users. 


\subsection{Stage 1: initial dataset}
\label{sec:stage_1}

Taking inspiration from the Qulac dataset~\cite{AliannejadiSigir19}~\footnote{Qulac is based on the TREC Web Track 2009-2012}, we have crowdsourced a new dataset to study clarifying questions that is suitable for conversational settings. Namely, the collected dataset consists of:
\begin{itemize}
    \item \textbf{User Request:} an initial user request in the conversational form, e.g. \emph{What is Fickle Creek Farm}, with a label reflects if clarification is needed ranged from 1 to 4;
    \item \textbf{Clarification questions:} a set of possible clarifying questions, e.g. \emph{do you want to know the location of fickle creek farm};
    \item \textbf{User Answers:} each questions is supplied with a user answer, e.g. \emph{no i want to find out where can i purchase fickle creek farm products}.
\end{itemize}

For training, the collected dataset is split into training (70\%) and validation (30\%) sets.  For testing, the participants are supplied with: (1)~ a set of user requests in conversational form and (2)~a set  a set of questions (i.e., question bank) which contains all the questions that we have collected for the collection. Therefore to answer our research questions in Section~\ref{sec:intro} we suggest the following two tasks:
\begin{itemize}
    \item To answer \textbf{RQ1}: Given a user request, return a score from $1$ to $4$ indicating the necessity of asking clarifying questions.
    \item To answer \textbf{RQ2}: Given a user request which needs clarification, return the most suitable clarifying question. 
\end{itemize}
Table~\ref{tab:stat_train} provides statistics for the train set.

\begin{table}[]
\label{tab:stat_train}
    \centering
    \caption{Statistics of SCAI Challenge Data - Train.}
    \begin{tabular}{ll}
        \toprule
         \# topics & 237 \\
         \# faceted topics & 141\\
         \# ambiguous topics & 57\\
         \# single topics & 39\\
         \midrule
         \# facets & 891\\
         \# informational facets & 577\\
         \# navigational facets & 185\\
         \midrule
         \# questions & 3,304\\
         \# question-answer pairs & 11,489\\
         Average terms per question & 9.70 $\pm$ 2.64\\
         Average terms per answer & 7.68 $\pm$ 4.75\\
         \bottomrule
    \end{tabular}
    \label{tab:stats_train}
    \vspace{-0.5cm}
\end{table}


As system automatic evaluation metrics we use MRR, P@[1,3,5,10,20], nDCG@[1,3,5,20].
These metrics are computed as follows: a selected clarifying question, together with its corresponding answer are added to the original request.
The updated query is then used to retrieve (or re-rank) documents from the collection. The quality of a question is then evaluated by taking into account how much the question and its answer affect the performance of document retrieval. 
Models are also evaluated in how well they are able to rank relevant questions higher than other questions in the question bank. For this task, that we call `question relevance', the models are evaluated in terms of Recall@[10,20,30]. Since the precision of models is evaluated in the document relevance task, here we focus only on recall.

The datasets and the scripts for automatic evaluation can be found at the following repository -- \url{https://github.com/aliannejadi/ClariQ}.


\subsection{Stage 2: human-in-the-loop}
\label{sec:stage_2}
At Stage~2 the participating systems are put in front of human users. The systems are rated on their overall performance.
At each dialog step, a system should give either a factual answer to the user's request or ask for a clarification question.
Therefore, the participants would need to:
\begin{itemize}
    \item ensure their system can answer simple user questions
    \item make their own decisions on when clarification might be appropriate
    \item provide clarification question whenever appropriate
    \item interpret user's answer to the clarifying question
\end{itemize}
The participants would need to strike a balance between asking too many questions
and providing irrelevant answers.

Note that the setup of this stage is quite different from the Stage~1.
Participating systems would likely need to operate as a \textit{generative} model,
rather than a \textit{retrieval} model. One option would be to cast the problem as generative from the beginning,
and solve the retrieval part of Stage~1, e.g.,
by ranking the offered candidates by their likelihood.

Alternatively, one may solve Stage 2 by retrieving a list of candidate answers
(e.g., by invoking Wikipedia API or the Chat Noir\footnote{\url{https://www.chatnoir.eu}} API that we describe above) and ranking them as in Stage~1.

\subsection{User-based evaluation}
We use the real humans to interact with the systems.
As a result, we get the following tuples \texttt{(Conversation history up until now, System's response, Ratings for the response)}
as well as overall rating for the interaction.
The users will rate each answer of the system on relevance and naturalness.

\begin{itemize}
    \item \textbf{Relevance.} Is the particular answer or clarifying question relevant to the user's information need (e.g., does it help ranking?).
    This can be used on the utterance level, independent of the dialog.
    \item \textbf{Naturalness.} Is the clarifying question natural \emph{in the context of the dialog}?
\end{itemize}

The separation between the two is to a large extent motivated by
the setup, where we assume the user has an information need, but
we don't want the system to approach it as a "yes/no" puzzle.
An example of relevant but unnatural conversation:
\begin{verbatim}
    User> Zurich zoo
    Syst> Do you want to know the opening hours?
    User> No.
    Syst> Do you want to know how many elephants does it have?
    User> You are reading my mind!

    %% The user does want to learn about the number of elephants there,
    %% but doesn't say so explicitly.
\end{verbatim}

\section*{Acknowledgements}
The challenge is organized as a joint effort by the University of Amsterdam, Microsoft, Google, University of Glasgow, and MIPT\@. We would like to thank Microsoft for their generous support of data annotation costs. We would also like to thank the Webis Group for giving us access to ChatNoir search API. Thanks to the crowd workers for their invaluable help in annotating ClariQ.
We would also like to stress that all content represents the opinion of the authors, which is not necessarily shared or endorsed by their respective employers and/or sponsors.

\bibliographystyle{apalike}
\bibliography{main}

\begin{thebibliography}{}

\bibitem[Aliannejadi et~al., 2019]{AliannejadiSigir19}
Aliannejadi, M., Zamani, H., Crestani, F., and Croft, W.~B. (2019).
\newblock Asking clarifying questions in open-domain information-seeking
  conversations.
\newblock In {\em International {ACM} {SIGIR} Conference on Research and
  Development in Information Retrieval (SIGIR)}, {SIGIR '19}.

\bibitem[Braslavski et~al., 2017]{braslavski2017you}
Braslavski, P., Savenkov, D., Agichtein, E., and Dubatovka, A. (2017).
\newblock What do you mean exactly? analyzing clarification questions in cqa.
\newblock In {\em Proceedings of the 2017 Conference on Conference Human
  Information Interaction and Retrieval}, pages 345--348.

\bibitem[Kwiatkowski et~al., 2019]{kwiatkowski2019natural}
Kwiatkowski, T., Palomaki, J., Redfield, O., Collins, M., Parikh, A., Alberti,
  C., Epstein, D., Polosukhin, I., Devlin, J., Lee, K., et~al. (2019).
\newblock Natural questions: a benchmark for question answering research.
\newblock {\em Transactions of the Association for Computational Linguistics},
  7:453--466.

\bibitem[Radlinski and Dumais, 2006]{radlinski2006improving}
Radlinski, F. and Dumais, S. (2006).
\newblock Improving personalized web search using result diversification.
\newblock In {\em Proceedings of the 29th annual international ACM SIGIR
  conference on Research and development in information retrieval}, pages
  691--692.

\bibitem[Rao and Daum{\'e}~III, 2018]{rao2018learning}
Rao, S. and Daum{\'e}~III, H. (2018).
\newblock Learning to ask good questions: Ranking clarification questions using
  neural expected value of perfect information.
\newblock {\em arXiv preprint arXiv:1805.04655}.

\bibitem[Rosset et~al., 2020]{rosset2020leading}
Rosset, C., Xiong, C., Song, X., Campos, D., Craswell, N., Tiwary, S., and
  Bennett, P. (2020).
\newblock Leading conversational search by suggesting useful questions.
\newblock In {\em The Web Conference 2020 (formerly WWW conference)}.

\bibitem[Trienes and Balog, 2019]{trienes2019identifying}
Trienes, J. and Balog, K. (2019).
\newblock Identifying unclear questions in community question answering
  websites.
\newblock In {\em European Conference on Information Retrieval}, pages
  276--289. Springer.

\bibitem[Zamani et~al., 2020a]{zamani2020generating}
Zamani, H., Dumais, S., Craswell, N., Bennett, P., and Lueck, G. (2020a).
\newblock Generating clarifying questions for information retrieval.
\newblock In {\em Proceedings of The Web Conference 2020}, pages 418--428.

\bibitem[Zamani et~al., 2020b]{zamani2020mimics}
Zamani, H., Lueck, G., Chen, E., Quispe, R., Luu, F., and Craswell, N. (2020b).
\newblock Mimics: A large-scale data collection for search clarification.
\newblock {\em arXiv preprint arXiv:2006.10174}.

\end{thebibliography}

\end{document}